%% file: main.tex
\documentclass[10pt]{article}

\usepackage[margin=1in]{geometry}
\usepackage[T1]{fontenc}
\usepackage[utf8]{inputenc}
\usepackage{mathptmx}
\usepackage{courier}
\usepackage{microtype}
\usepackage{graphicx}
\usepackage{booktabs}
\usepackage{multirow}
\usepackage{array}
\usepackage{tabularx}
\usepackage{amsmath}
\usepackage{amssymb}
\usepackage{xcolor}
\usepackage[numbers,sort&compress]{natbib}
\usepackage[font=small,labelfont=bf]{caption}
\usepackage[colorlinks=true,linkcolor=blue!60!black,citecolor=blue!60!black,urlcolor=blue!60!black]{hyperref}

\newcommand{\bench}{DAYJOB}
\newcommand{\hcname}{\bench: Healthcare}
\newcommand{\finname}{\bench: Finance}
\newcolumntype{L}[1]{>{\raggedright\arraybackslash}p{#1}}
\input{generated/numbers.tex}


\hypersetup{
  pdftitle={DAYJOB: A Benchmark for Long-Horizon Professional Work},
  pdfauthor={Surge AI}
}

\title{\bench: A Benchmark for Long-Horizon Professional Work}

\author{Stephanie Finley, Liudas Panavas, Thomas Mikkelson, Cam Hinton, Stacey Ganss,\\
Bradley Monton, Emily Kendall, Michelle Spradlin, Lydia Bye, Michael O'Brien,\\
Lauren Ylvisaker, Derek Ray, Suhaas Garre, Sushant Mehta, Edwin Chen\\[4pt]
Surge AI\thanks{Correspondence to: \texttt{benchmarks@surgehq.ai}. Tasks: \url{https://huggingface.co/datasets/surgeai/DAYJOB-healthcare} and \url{https://huggingface.co/datasets/surgeai/DAYJOB-finance}. Harness: \url{https://github.com/surge-ai/dayjob}. Leaderboards: \url{https://surgehq.ai/benchmarks/dayjob-healthcare} and \url{https://surgehq.ai/benchmarks/dayjob-finance}.}
}
\date{}

\begin{document}
\maketitle

\begin{abstract}
Professional work often starts with a brief request that leaves the professional to work out what is needed, which documents matter, and whether the request's premise holds. We introduce \bench{}, a benchmark of \allTasks{} tasks built by professionals in healthcare (\hcTasks{}) and finance (\finTasks{}). The tasks are estimated to take a professional \hcHoursMean{} hours on average in healthcare and \finHoursMean{} in finance. Each task is a containerized Harbor environment with an expert rubric of binary criteria (median \hcCritMedian{} and \finCritMedian{} per task) that an agentic judge applies to the delivered files, and an attempt passes only if it meets every criterion. Across \nConfigs{} model configurations from \nDevelopers{} developers, the strongest, Claude Opus 5.5, passes \hcTopScore\% of healthcare and \finTopScore\% of finance attempts, and the median configuration passes \hcMedianStrict\% and \finMedianStrict\%. In case studies, agents accept premises that the record contradicts and carry wrong inputs through otherwise consistent analyses. We release all healthcare tasks, \finStatsTasks{} of the \finTasks{} finance tasks, the evaluation harness, and the leaderboard.
\end{abstract}

\section{Introduction}
\label{sec:intro}

Requests at work assume shared context. Someone preparing a new client's first fund reports asks whether anything should stop the reports from going out today. An urgent-care clinician asks for a read on the labs before discharging a patient with ``textbook Bell's Palsy'' on prednisone. Neither request mentions the problem it contains. In the fund's reporting pack, an equity price quoted in South African cents on a Bloomberg screen was entered as rand, so a position worth about \$260{,}000 appears to be worth \$26 million. The patient's chart records right-sided neck pain after moving heavy furniture, slightly slurred speech, and a facial examination that does not document the forehead or eyelid closure, and the task's rubric requires ruling out an arterial dissection before any discharge. Both are \bench{} tasks (Section~\ref{sec:cases}).

GDPval~\citep{patwardhan2025gdpval} builds tasks from the work of experienced professionals and gives the model a detailed brief: prompts in its public gold subset average \gdpPromptMean{} words, and \gdpAtMostOneFilePct\% of its tasks include at most one reference file. GDPval also reports that GPT-5 performed worse when the prompts omitted where to find data and how to approach the problem. Agent benchmarks supply interactive environments, mostly for software, web, and desktop tasks~\citep{jimenez2024swebench,zhou2024webarena,xie2024osworld,xu2024theagentcompany}, and healthcare and finance benchmarks mostly test knowledge, question answering, or bounded tool use~\citep{jin2021medqa,arora2025healthbench,jiang2025medagentbench,chen2021finqa,islam2023financebench,bigeard2025financeagent}.

\bench{} pairs realistic requests of the kind a colleague would send with a workspace of the files they implicitly refer to, mostly PDFs, spreadsheets, and Word documents, and with a rubric of binary criteria that describes a competent deliverable. Rubrics credit actions the request leaves unsaid, such as recommending urgent transfer to a stroke-capable emergency department, and prohibit specific errors. Tasks run as containerized Harbor environments~\citep{harbor2026} with network access limited to the model provider's API. After the agent finishes, an agentic judge in the same container opens the delivered files, recomputes values where needed, and returns a verdict with evidence for every criterion. An attempt passes only if every criterion is met.

We evaluate \nConfigs{} configurations from \nDevelopers{} developers with five attempts per task. \hcTopModel{} leads both domains, passing \hcTopScore\% of healthcare and \finTopScore\% of finance attempts. The median configuration passes \hcMedianStrict\% and \finMedianStrict\%, and \bothBelowFive{} configurations score below 5\% in both domains. We release all \hcTasks{} healthcare tasks, \finStatsTasks{} of the \finTasks{} finance tasks, the harness, and the leaderboard.

\section{Related work}
\label{sec:related}

\paragraph{Evaluations of economically valuable work.}
GDPval~\citep{patwardhan2025gdpval} spans 1{,}320 tasks from 44 occupations in the nine sectors that contribute most to U.S.\ GDP and releases a 220-task gold subset, 89\% of which its expert graders rated as well specified. Experts compare model deliverables with human ones in blinded pairwise judgments. APEX~\citep{vidgen2025apex} grades responses to 400 held-out cases, 100 each from investment banking, management consulting, big law, and primary care, against expert rubrics. On the Remote Labor Index~\citep{mazeika2025rli}, a collection of real freelance projects, the best agent achieves an automation rate of 2.5\%, and SWE-Lancer~\citep{miserendino2025swelancer} values freelance software tasks by their real payouts. TheAgentCompany~\citep{xu2024theagentcompany} simulates a software company with internal web services and simulated colleagues. \bench{} differs in giving the agent a brief request and a large workspace, and in requiring a task-specific rubric to be met in full.

\paragraph{Agent benchmarks.}
Interactive benchmarks measure goal completion in software engineering~\citep{jimenez2024swebench}, web navigation~\citep{zhou2024webarena,drouin2024workarena}, operating systems~\citep{xie2024osworld}, tool use with simulated users~\citep{yao2024taubench}, general assistance~\citep{mialon2023gaia}, and long-horizon tasks across many applications~\citep{li2026toolathlon}. Enterprise benchmarks place agents in CRM systems~\citep{huang2025crmarenapro} and a simulated customer-support organization~\citep{mehta2026corecraft}, or test adherence to long company policies~\citep{panavas2026handbook}. Time-horizon measurements relate agent success to how long tasks take skilled humans~\citep{kwa2025timehorizon}. \bench{} gives the agent only general tools: a terminal, a file editor, and a task tracker.

\paragraph{Healthcare and finance.}
Medical evaluation has moved from licensing-exam questions~\citep{jin2021medqa} to physician-written rubrics for 5{,}000 conversations~\citep{arora2025healthbench}, simulated clinical encounters~\citep{schmidgall2024agentclinic}, and 300 agent tasks in a FHIR electronic health record~\citep{jiang2025medagentbench}. Financial evaluation spans numerical reasoning over reports~\citep{chen2021finqa}, question answering over company filings~\citep{islam2023financebench}, and research questions answered with search and EDGAR tools~\citep{bigeard2025financeagent}. \bench{} tasks end in work products, such as chart notes, appeal letters, credit memos, and audit workpapers, built from heterogeneous records, and many healthcare tasks concern payer, pharmacy, compliance, and operations work.

\paragraph{Rubrics and model judges.}
Rubric-based grading with model judges is common for open-ended work~\citep{arora2025healthbench,starace2025paperbench,gunjal2025rubrics,garre2026gdppdf,mehta2026complexconstraints}. Strong model judges can match the agreement between human raters on open-ended conversation~\citep{zheng2023judging}. Agent-as-a-Judge~\citep{zhuge2024agentjudge} lets the judge act in the environment to verify the work, as the \bench{} judge does.

\section{The \bench{} benchmark}
\label{sec:benchmark}

\subsection{Task definition}
\label{sec:task}

A task $i$ consists of a request $q_i$, an environment $E_i$, and a rubric $R_i=\{c_{i1},\dots,c_{iK_i}\}$ of $K_i$ binary criteria. The environment is a container image whose working directory holds the task's input files. The agent receives an instruction made of a preamble shared by all tasks followed by $q_i$, works inside $E_i$ without human interaction, and ends with a final message. Its deliverables are the files it creates or modifies and, when the task asks for a written answer, the final message. A judge inspects the final workspace and the agent's trajectory and assigns a verdict $g_{ij}\in\{0,1\}$ to every criterion. The attempt's reward is the fraction of criteria met, $r=\frac{1}{K_i}\sum_{j=1}^{K_i} g_{ij}$; the attempt passes strictly if $r=1$ and passes at threshold $\tau$ if $r\ge\tau$.

\subsection{Design principles}
\label{sec:design}

\paragraph{Requests.}
The shortest healthcare request reads, in full, ``Please draft a brief document for this case for the upcoming M\&M.'' Others name the deliverable and its format and leave the rest to the files: ``Take a look at the IVIG denial. Find the root cause of the denial and draft the appeal you'd recommend. Give me both in .docx''. Longer requests add background or several deliverables. One asks for a physician-facing patient summary of at most 3{,}000 words, a family-facing version, and letters to the patient's primary care pediatrician and to the subspecialists named in the chart. The preamble shared by all tasks (in each task's \nolinkurl{instruction.md}) tells the agent to infer its role, workplace, and deliverables from the files and to produce its best-supported output without asking for clarification, and the solvability review (Section~\ref{sec:construction}) checks that the files suffice.

\paragraph{Workspaces.}
Input files are the documents the work draws on: charts and encounter notes, claims exports, laboratory reports, policies, vendor price files, general-ledger extracts, loan agreements, and email. Most are PDFs (\hcFmtPDFPct\% in healthcare and \finFmtPDFPct\% in finance), and most of the rest are spreadsheets and Word documents. Relevant and irrelevant material sit side by side: in the urgent-care task, seven of the nine PDFs are administrative forms, and the chart in a skilled-nursing task includes maintenance and fire-drill records. Workspaces also contain stale or conflicting sources, such as a 2019 site evaluation in a 2026 expansion proposal, or a bank memo's 3.25\% borrowing rate that the chief financial officer's later chat message, supplied as a screenshot, raises to 5.25\%. Some evidence exists only in images (Section~\ref{sec:example}).

\paragraph{Rubrics.}
Criteria are binary statements about the delivered work. Most check substance, often a value within a stated tolerance. Others check actions the request never mentions or the internal consistency of a set of deliverables. Prohibitions fail an attempt that makes a specific, plausible error, such as flagging a correctly priced post-split equity as a release blocker, recommending surgical debridement of a wound whose presentation suggests pyoderma gangrenosum, or fabricating results for tests that were never performed. We count a criterion as a prohibition when its main clause is negated, as in criteria that begin ``Does not''. By this count, all \hcProhibTasks{} healthcare rubrics and \finProhibTasks{} of the \finStatsTasks{} public finance rubrics contain at least one prohibition, and prohibitions make up \hcProhibPct\% and \finProhibPct\% of criteria.

\subsection{An example task}
\label{sec:example}

Table~\ref{tab:example} excerpts a finance task whose request asks whether anything should stop the first reports for a new client's fund. The workspace holds 13 files: the firm's pricing-source hierarchy and approved broker register, a broker quote recap, five price and exchange-rate files from four sources, three reports exported from the fund-accounting system, and two emails. The ten equity prices were keyed by hand from Bloomberg screens attached to one of the emails. The screen for Nerovexa quotes the price in ZAr, South African cents, while the email's typed price table gives the currency as ZAR, and the price was entered as rand. The 20 criteria require finding and quantifying this error and two pricing-hierarchy exceptions, withholding the reports until the Nerovexa price is corrected, recognizing the manual keying as a control weakness, and not raising false alarms about prices that follow policy.

\begin{table}[t]
\centering
\small
\caption{Excerpt of a \finname{} task (\texttt{finance\_9148d239}): six of its 20 criteria, abridged.}
\label{tab:example}
\begin{tabular}{@{}L{0.97\linewidth}@{}}
\toprule
\multicolumn{1}{@{}L{0.97\linewidth}@{}}{\textbf{Request.} \emph{``Caldervane Emerging Markets Fund is a new client's first fund with us and we are delivering our first reports for it today for COB June 30th. The client wants the initial reports as soon as they are available, so they can review these BAU deliverables in detail. Go through the attached pack and let me know if there are any issues stopping me sending the reports to the client as they are. Write it up as a short .md memo.''}} \\
\midrule
\textbf{Criteria} \\
\midrule
Identifies that the Bloomberg screen quotes the Nerovexa price of 261.200 in South African cents rather than rand. \\
States that the Nerovexa position is overstated by about \$25.7 million (accept \$25.5m--\$25.9m). \\
Recommends not sending the reports until the Nerovexa position is corrected; a hold that lifts once the quotation basis is confirmed does not satisfy the criterion. \\
Identifies that the Kingdom of Saudi Arabia 4.500\% 2046 bond was priced from BVAL when ICE supplied a current evaluation. \\
Does not call the Dravena position quantity or its 1{,}242.000 CZK price incorrect, and does not treat the stock split as a reason to withhold the reports. \\
Identifies that the 30 June equity prices were keyed manually because the equity price schedule was not assigned to the new portfolio, and treats this as a control weakness. \\
\bottomrule
\end{tabular}
\end{table}

\subsection{Composition}
\label{sec:stats}

\begin{table}[t]
\centering
\small
\caption{\bench{} compared with the GDPval gold subset. Counts come from each benchmark's task files (request text, input workspace, and rubric). Finance statistics describe the \finStatsTasks{} public tasks. $^\ddagger$Healthcare hours average the task creators' range estimates at range midpoints, with ``more than 40 hours'' counted as 40; the GDPval figure is the mean validated self-reported completion time on the gold subset (404 minutes) reported by \citet{patwardhan2025gdpval}. $^\S$Average reported on the finance dataset card~\citep{surge2026dayjobhf}.}
\label{tab:dataset}
\begin{tabular}{@{}lccc@{}}
\toprule
 & \hcname & \finname & GDPval gold subset \\
\midrule
\input{generated/tab_dataset.tex} 
Rubric format & binary criteria & binary criteria & weighted items \\
Grading & \multicolumn{2}{c}{agentic model judge, all criteria required} & pairwise expert comparison \\
\bottomrule
\end{tabular}
\end{table}

\begin{figure}[t]
\centering
\includegraphics[width=\linewidth]{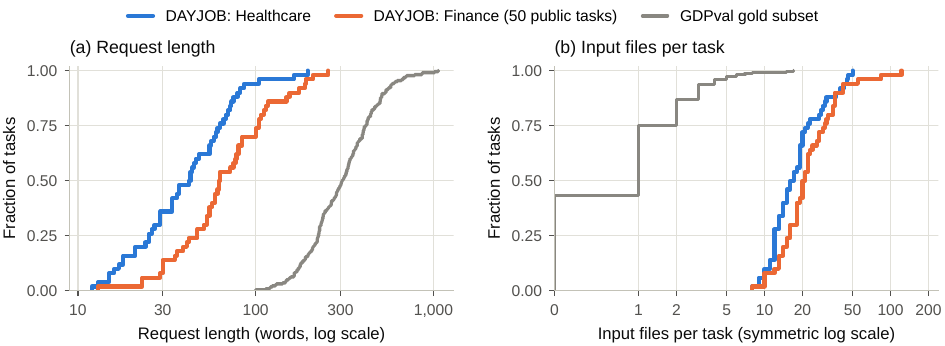}
\caption{Cumulative distributions of (a) request length and (b) the number of input files per task. The shortest GDPval request (\gdpPromptMin{} words) is longer than \hcBelowGdpMinPct\% of healthcare requests and \finBelowGdpMinPct\% of public finance requests, and \gdpNoFilePct\% of GDPval tasks have no input file.}
\label{fig:requests}
\end{figure}

Table~\ref{tab:dataset} and Figure~\ref{fig:requests} compare \bench{} with the GDPval gold subset. GDPval requests are \ratioPromptHC{} times as long on average as healthcare requests and \ratioPromptFIN{} times as long as finance requests, while \bench{} workspaces hold \ratioFilesHC{} and \ratioFilesFIN{} times as many files, including files that are irrelevant by design. Most tasks have at least ten input files (\hcFilesTenPlusPct\% in healthcare and \finFilesTenPlusPct\% in finance). Rubric sizes are comparable, with medians of \hcCritMedian{} and \finCritMedian{} criteria against \gdpCritMedian{} GDPval rubric items, but every \bench{} criterion must be met.

Task creators estimated the time a professional would need for each healthcare task, choosing one of eight ranges from 1--2 hours to more than 40 hours (Table~\ref{tab:dataset}). The median task falls in the \hcHoursMedianRange{} hour range, and \hcHoursTwentyPlusPct\% of tasks are estimated at 20 hours or more.

\paragraph{Coverage.}
Healthcare tasks include clinical decision support (discharge decisions, ICU admission notes, workup plans, medication reconciliation), pharmacy dispensing and parenteral nutrition, revenue-cycle and payer work (denial appeals, prior authorization, inpatient coding, contract negotiation), quality, safety, and compliance reviews (morbidity and mortality, a HIPAA breach audit, a trauma registry), and allied-health documentation. Finance tasks include valuation and corporate finance (discounted cash flow, accretion and dilution, capital structure), credit (term-loan requests, watchlist memos, covenant compliance), real estate and project finance (acquisition underwriting, construction draws), fund operations (net asset value, pricing review, rebalancing), treasury (cash forecasts, hedge close), and close and audit work (lease accounting, tie-out workpapers).

\subsection{Construction, selection, and release}
\label{sec:construction}

Tasks were created by professionals with firsthand experience of the work, among them a senior portfolio manager in institutional asset management, a finance professional who has structured debt for infrastructure and renewable-energy projects, and a pediatric endocrinologist~\citep{surge2026dayjobblog}. Every task went through three layers of review for correctness, completeness, and solvability.

All \hcTasks{} healthcare tasks and \finStatsTasks{} of the \finTasks{} finance tasks are public on Hugging Face under the MIT license~\citep{surge2026dayjobhf}, and the remaining finance tasks are available on request~\citep{surge2026dayjobblog}. The harness is released under Apache 2.0, and the scripts behind every statistic, table, and figure in this paper are included as ancillary files. Each task directory holds the instruction, input files, rubric, and judge prompt, together with a benchmark-specific canary string and a SHA-256 hash of its file tree, and the dataset cards ask that tasks and their derivatives be excluded from training and from checkpoint selection.

\section{Evaluation protocol}
\label{sec:protocol}

\subsection{Harness}
\label{sec:harness}

Every attempt runs the released harness: an agent loop built on the OpenHands Software Agent SDK~\citep{wang2025openhandssdk,wang2025openhands} (version 1.43.1) that executes inside the task container through a Harbor adapter (Harbor 0.22.0). The agent has a terminal, a file editor, and a task tracker. Models with vision support can view image files through the editor, and images embedded in PDFs must first be extracted or rendered. The SDK's finish and think tools are disabled, so an attempt ends at the first reply that contains text and no tool call. The instruction is the first user message, and the system prompt is the SDK's default. When the history exceeds 240 events or the model's input limit, a summarizing condenser that calls the evaluated model replaces older events with a summary. Each attempt is limited to 1{,}000 agent turns, \agentTimeoutHours{} hours, and 600 seconds per model call. The container has \taskCPUs{} CPUs and \taskMemGB{}~GB of memory and provides Python 3.12 with document libraries (openpyxl, python-docx, pdfplumber, PyMuPDF, and others), LibreOffice, Poppler, Tesseract OCR, and Ghostscript. Network egress is limited to the model provider's API host, so web search, package installation, and lookup of the released tasks are unavailable.

\subsection{Models and attempts}
\label{sec:models}

We evaluate the \nConfigs{} configurations from \nDevelopers{} developers on the launch leaderboard of September 2026~\citep{surge2026dayjobleaderboard} (Table~\ref{tab:leaderboard}). Parenthesized labels give the reasoning setting passed to the provider, and ``adaptive/max'' denotes adaptive thinking at maximum effort. Each configuration attempts every task five times, and Harbor reruns an attempt up to three times after a provider rate-limit error. Attempts that end in an error, such as a time-out or a provider failure, are excluded from their task's mean.

\subsection{Grading}
\label{sec:grading}

After the agent phase, the verifier runs an agentic judge in the same container: the OpenCode agent~\citep{opencode2026} (version 1.18.31), run by rewardkit 0.1.7 with Claude Opus 4.8 as its model. In a single session the judge receives its instructions (\nolinkurl{tests/prompt_template.md} in each task), the request, the names of the input files, and all criteria. It can open any file, run code, and read the agent's trajectory, and it returns a binary verdict with reasoning for each criterion. The instructions tell it to grade the files the agent created or modified and, when the task calls for a written answer, the final message; to cite the file and content behind each verdict; to recompute values instead of trusting a formula or script; and to inspect rendered outputs. Claims in the final message about what a file contains are not evidence. The judge may take \judgeSecondsPerCriterion{} seconds per criterion, with a minimum of \judgeMinMinutes{} minutes.

\subsection{Metrics}
\label{sec:metrics}

For configuration $m$, domain $d$ with task set $\mathcal{T}_d$, and the scored attempts $A_{mi}$ of task $i$ (at most five), pass@1 at threshold $\tau$ is
\begin{equation}
\mathrm{pass@1}_{\tau}(m,d)\;=\;\frac{1}{|\mathcal{T}_d|}\sum_{i\in\mathcal{T}_d}\frac{1}{|A_{mi}|}\sum_{a\in A_{mi}}\mathbf{1}\!\left[r_{mia}\ge\tau\right],
\label{eq:pass}
\end{equation}
where $r_{mia}$ is the reward of attempt $a$. The headline metric is \emph{strict pass@1}, $\tau=1$. We also report $\tau\in\{0.95, 0.90\}$ and pass@1~($N{-}1$), which accepts attempts that fail at most one criterion. At $\tau=0.95$ an attempt may miss $\lfloor 0.05K_i\rfloor$ criteria, between \hcMissFiveMin{} and \hcMissFiveMax{} in healthcare. Tasks are weighted equally within a domain.

Criteria are not weighted by importance, so a lower threshold accepts attempts that miss a decisive one. In the task of Table~\ref{tab:example}, an attempt that finds the \$25.7 million overstatement but lifts its hold once the quotation basis is confirmed can fail only the release criterion, the third in the table, and still pass at 95\%.

\section{Results}
\label{sec:results}

\subsection{Main results}
\label{sec:main}

\begin{table}[t]
\centering
\small
\caption{Launch leaderboard (September 2026): strict pass@1 (\%) and rank on \hcTasks{} healthcare and \finTasks{} finance tasks, five attempts per task. Rows are ordered by the mean of the two scores, and equal displayed scores share a rank.}
\label{tab:leaderboard}
\setlength{\tabcolsep}{4pt}
\begin{tabular}{@{}llcccc@{}}
\toprule
 & & \multicolumn{2}{c}{\hcname} & \multicolumn{2}{c}{\finname} \\
\cmidrule(lr){3-4}\cmidrule(l){5-6}
Configuration & Developer & Strict & Rank & Strict & Rank \\
\midrule
\input{generated/tab_leaderboard.tex} 
\bottomrule
\end{tabular}
\end{table}

Table~\ref{tab:leaderboard} lists strict pass@1 for all \nConfigs{} configurations, and Figure~\ref{fig:leaderboard} (Appendix~\ref{app:full}) plots it. \hcTopModel{} leads both domains, with \hcTopScore\% in healthcare and \finTopScore\% in finance. In healthcare the next configuration, \hcSecondModel{}, scores \hcSecondScore\%, \hcTopGap{} points lower. In finance the top three configurations lie within \finTopThreeSpread{} points (\finTopScore\%, \finSecondScore\%, and \finThirdScore\%). The median configuration scores \hcMedianStrict\% in healthcare and \finMedianStrict\% in finance. In healthcare \hcZero{} configurations never pass an attempt, and in finance \finZero{} never do.

The two domains order configurations similarly: among the \nNonzeroEither{} configurations with a nonzero score in either domain, the Spearman correlation between domains is $\rho=\rhoDomainsNonzero$. Scores are higher in finance for \finHigherCount{} configurations and higher in healthcare for \hcHigherCount{}, \hcHigherNames{}.

\subsection{Relaxed thresholds}
\label{sec:threshold}

\begin{figure}[!tb]
\centering
\includegraphics[width=\linewidth]{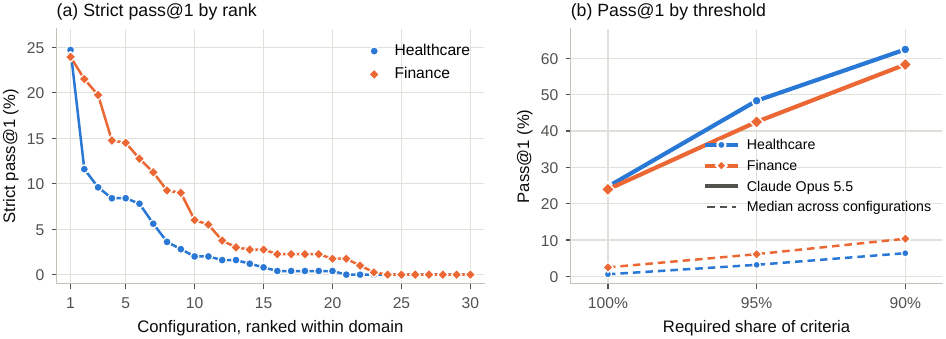}
\caption{(a) Strict pass@1 of the \nConfigs{} configurations, sorted within each domain. (b) Pass@1 as the required share of criteria is relaxed, for the leader and for the median across configurations at each threshold.}
\label{fig:threshold}
\end{figure}

In healthcare \hcAboveFive{} configurations reach 5\% strict pass@1, and in finance \finAboveFive{} (Figure~\ref{fig:threshold}a). Counting attempts that meet 95\% of criteria multiplies the scores of the top five configurations by \hcLiftTopFive{} in healthcare and \finLiftTopFive{} in finance on average. At 90\% the leader reaches \hcTopNinety\% and \finTopNinety\%, and the median configuration \hcMedianNinety\% and \finMedianNinety\% (Figure~\ref{fig:threshold}b; all values are in the ancillary data). Under pass@1~($N{-}1$) the leader reaches \hcTopNOne\% and \finTopNOne\%. The leader is first at every threshold, and rankings at 95\% agree closely with strict rankings (Kendall $\tau_b=\hcTauStrictNinetyFive$ in healthcare and \finTauStrictNinetyFive{} in finance).

\subsection{Cost and tokens}
\label{sec:cost}

\begin{figure}[!tb]
\centering
\includegraphics[width=\linewidth]{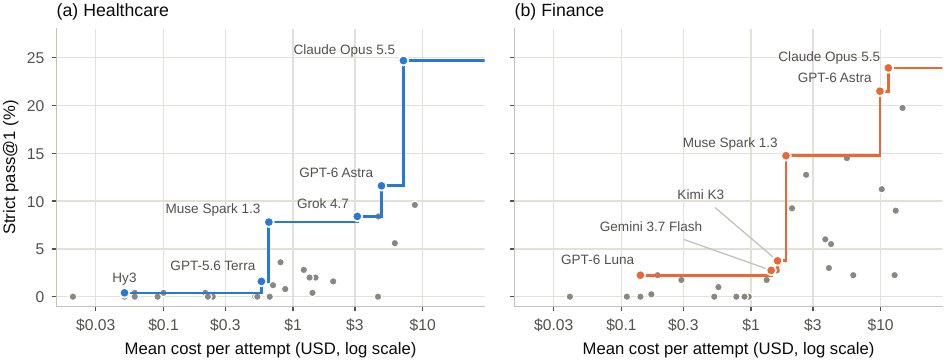}
\caption{Strict pass@1 against mean cost per attempt. Colored points are the configurations on the cost frontier, and the step line gives the best strict score available at or below each cost; gray points are the other configurations. Costs exclude the judge.}
\label{fig:cost}
\end{figure}

Mean cost per attempt ranges from \$\hcCostMin{} to \$\hcCostMax{} in healthcare and from \$\finCostMin{} to \$\finCostMax{} in finance (Figure~\ref{fig:cost}). It covers the agent's and the condenser's model calls at litellm's prices, or at provider billing where litellm has no price, and excludes the judge. In both domains the cost frontier ends at \hcTopModel{}, at \$\hcTopCost{} per healthcare attempt and \$\finTopCost{} per finance attempt, and includes GPT-6 Astra (max) and Muse Spark~1.3 (max) at lower cost. The most expensive configuration, \hcCostMaxModel{}, is not on it. The leader processes \hcTopTokensM{} million tokens per healthcare attempt and \finTopTokensM{} million per finance attempt, \hcTopCachedPct\% and \finTopCachedPct\% of them cached input.

\section{Case studies}
\label{sec:cases}

The failed attempts below are published on the leaderboard pages with their requests, response excerpts, and rubric verdicts~\citep{surge2026dayjobleaderboard}.

\paragraph{Unchallenged premises.}
In the urgent-care task (\texttt{healthcare\_a6b300f3}), GPT-5.6 Sol (max) read the labs and found it reasonable to start prednisone and discharge the patient. Its reasoning noted that facial weakness with neck pain could have a vascular cause, then concluded, ``I shouldn't question the diagnosis unless necessary''~\citep{surge2026dayjobblog}. The rubric requires the note to challenge the diagnosis, to withhold prednisone until cervicocephalic artery dissection or another acute ischemic event is excluded, and to recommend urgent transfer to a stroke-capable emergency department. An attempt by a Grok model on the same record connected the incomplete examination and the neck pain to a possible arterial injury and recommended holding the discharge for emergency evaluation and vascular imaging~\citep{surge2026dayjobblog}. In a second task, the request asks for an SBAR transfer summary for a patient going to surgery for debridement of an amputation stump. The chart describes two small papules at the incision that opened into an enlarging ulcer with dusky, irregular edges, 7/10 pain at dressing changes, and no response to two courses of oral antibiotics. Grok~4.6 (xhigh) gathered these facts and recommended aggressive surgical debridement. The rubric expects the summary to raise pyoderma gangrenosum, which surgical trauma can worsen, and to advise holding debridement until a biopsy or specialist evaluation.

\paragraph{Wrong inputs in consistent analyses.}
In the fund-reporting task of Table~\ref{tab:example}, GPT-5.6 Sol (max) found both pricing-hierarchy exceptions, which net to a \$132{,}000 understatement, and recommended holding the reports. It did not question the Nerovexa price, and its final message stated ``No other release blocker was identified,'' leaving the \$25.7 million overstatement in place. Asked whether a furniture retailer should fund a 2026 expansion and how to finance it, Grok~4.6 (xhigh) built discounted cash flows for three candidate sites and recommended opening one store with new senior debt, at a pro-forma leverage of about 2.66$\times$ against a 3.00$\times$ covenant. The existing-store forecasts it relied on add returns and allowances to gross sales for five stores instead of subtracting them, which inflates the earnings in its leverage test. With the signs corrected, the debt-funded buildout breaches the covenant. Asked for an annual cost comparison of in-house and outsourced check processing for 2021--2025, Qwen~3.8 Max (xhigh) counted electronic disbursements such as payroll and tax payments as printed checks, and so used 2{,}838 checks for 2025 against a true volume of 138. It recommended outsourcing, which at the true 2025 volume costs more than processing checks in-house.

\section{Conclusion}
\label{sec:conclusion}

\bench{} tests whether agents can turn the kind of request a colleague would send in realistic workspaces into deliverables that meet an expert rubric in full. Its \allTasks{} tasks come from the work of healthcare and finance professionals and test whether an agent can decide what the task is, find the facts that matter, and recognize when a request's premise is wrong.

The strongest of \nConfigs{} model configurations passes \hcTopScore\% of healthcare and \finTopScore\% of finance attempts, the median configuration passes \hcMedianStrict\% and \finMedianStrict\%, and \bothBelowFive{} configurations score below 5\% in both domains. In the case studies, agents accept diagnoses and treatment plans that the record contradicts and build consistent analyses on wrong inputs. Because the judge returns a verdict with evidence for every criterion, each failed attempt shows which fact or decision the agent missed.

We release all healthcare tasks and \finStatsTasks{} of the \finTasks{} finance tasks with the evaluation harness and the leaderboard, and will extend \bench{} to further professional domains in the same format.

\bibliographystyle{abbrvnat}
\bibliography{references}

\clearpage
\appendix

\section{Strict pass@1 by configuration}
\label{app:full}

\begin{figure}[h]
\centering
\includegraphics[width=\linewidth]{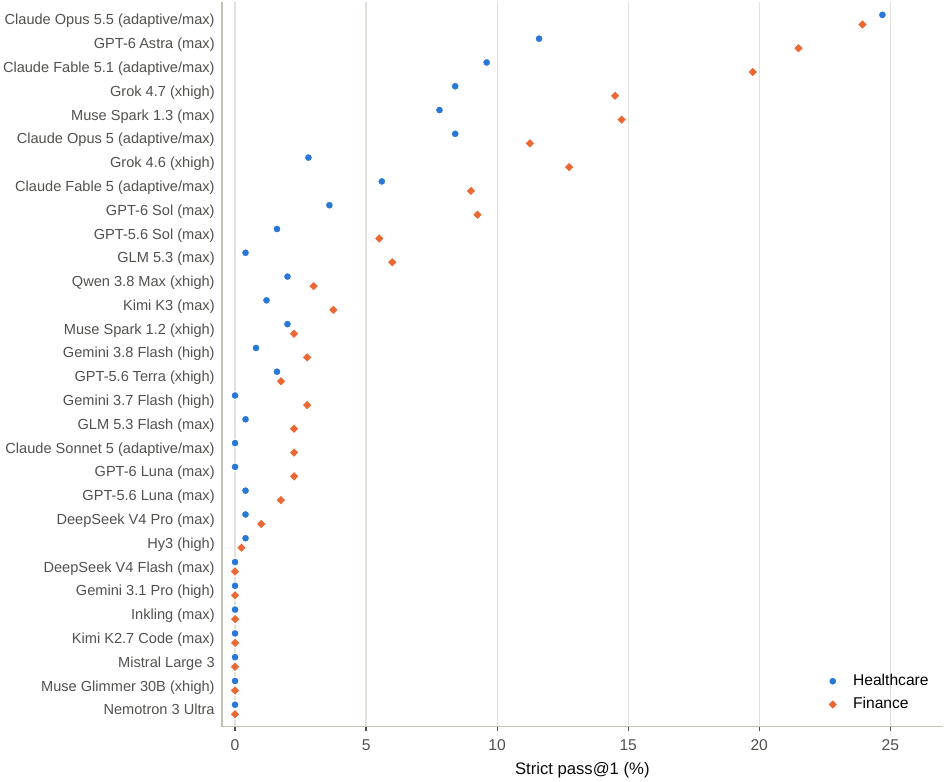}
\caption{Strict pass@1 for all \nConfigs{} configurations, in the order of Table~\ref{tab:leaderboard}.}
\label{fig:leaderboard}
\end{figure}

\end{document}

%% file: generated/numbers.tex
\newcommand{\hcTasks}{50}
\newcommand{\finTasks}{80}
\newcommand{\allTasks}{130}
\newcommand{\finStatsTasks}{50}
\newcommand{\hcPromptMean}{48.9}
\newcommand{\hcPromptMedian}{42.5}
\newcommand{\hcFilesMean}{19.8}
\newcommand{\hcFilesMedian}{16.5}
\newcommand{\hcFilesMin}{8}
\newcommand{\hcFilesMax}{50}
\newcommand{\hcCritMean}{55.7}
\newcommand{\hcCritMedian}{47.5}
\newcommand{\hcFilesTenPlusPct}{94}
\newcommand{\hcFmtPDFPct}{80}
\newcommand{\hcProhibTasks}{50}
\newcommand{\hcProhibPct}{16}
\newcommand{\finPromptMean}{80.7}
\newcommand{\finPromptMedian}{62.5}
\newcommand{\finFilesMean}{25.7}
\newcommand{\finFilesMedian}{20.5}
\newcommand{\finFilesMin}{8}
\newcommand{\finFilesMax}{122}
\newcommand{\finCritMean}{66.1}
\newcommand{\finCritMedian}{57.5}
\newcommand{\finFilesTenPlusPct}{98}
\newcommand{\finFmtPDFPct}{70}
\newcommand{\finProhibTasks}{43}
\newcommand{\finProhibPct}{8}
\newcommand{\hcMissFiveMin}{1}
\newcommand{\hcMissFiveMax}{6}
\newcommand{\agentTimeoutHours}{6}
\newcommand{\taskCPUs}{2}
\newcommand{\taskMemGB}{4}
\newcommand{\judgeSecondsPerCriterion}{90}
\newcommand{\judgeMinMinutes}{40}
\newcommand{\hcHoursMean}{13.6}
\newcommand{\finHoursMean}{16.6}
\newcommand{\hcHoursMedianRange}{5--10}
\newcommand{\hcHoursTwentyPlusPct}{32}
\newcommand{\gdpTasks}{220}
\newcommand{\gdpPromptMean}{337.2}
\newcommand{\gdpPromptMedian}{307}
\newcommand{\gdpPromptMin}{100}
\newcommand{\gdpFilesMean}{1.19}
\newcommand{\gdpFilesMedian}{1}
\newcommand{\gdpFilesMax}{17}
\newcommand{\gdpNoFilePct}{43}
\newcommand{\gdpAtMostOneFilePct}{75}
\newcommand{\gdpCritMean}{47.5}
\newcommand{\gdpCritMedian}{47}
\newcommand{\hcBelowGdpMinPct}{94}
\newcommand{\finBelowGdpMinPct}{70}
\newcommand{\ratioPromptHC}{6.9}
\newcommand{\ratioPromptFIN}{4.2}
\newcommand{\ratioFilesHC}{17}
\newcommand{\ratioFilesFIN}{22}
\newcommand{\nConfigs}{30}
\newcommand{\nDevelopers}{13}
\newcommand{\hcTopScore}{24.7}
\newcommand{\hcSecondScore}{11.6}
\newcommand{\hcTopNinety}{62.4}
\newcommand{\hcTopNOne}{37.8}
\newcommand{\hcZero}{10}
\newcommand{\hcMedianStrict}{0.6}
\newcommand{\hcMedianNinety}{6.4}
\newcommand{\hcAboveFive}{7}
\newcommand{\hcLiftTopFive}{2.7}
\newcommand{\hcTauStrictNinetyFive}{0.85}
\newcommand{\hcTopCost}{7.10}
\newcommand{\hcTopTokensM}{9.8}
\newcommand{\hcTopCachedPct}{94}
\newcommand{\hcCostMin}{0.02}
\newcommand{\hcCostMax}{8.70}
\newcommand{\finTopScore}{23.9}
\newcommand{\finSecondScore}{21.5}
\newcommand{\finTopNinety}{58.3}
\newcommand{\finTopNOne}{32.9}
\newcommand{\finZero}{7}
\newcommand{\finMedianStrict}{2.5}
\newcommand{\finMedianNinety}{10.4}
\newcommand{\finAboveFive}{11}
\newcommand{\finLiftTopFive}{1.8}
\newcommand{\finTauStrictNinetyFive}{0.92}
\newcommand{\finTopCost}{11.47}
\newcommand{\finTopTokensM}{17.8}
\newcommand{\finTopCachedPct}{95}
\newcommand{\finCostMin}{0.04}
\newcommand{\finCostMax}{14.76}
\newcommand{\hcTopModel}{Claude Opus 5.5 (adaptive/max)}
\newcommand{\hcSecondModel}{GPT-6 Astra (max)}
\newcommand{\hcTopGap}{13.1}
\newcommand{\finThirdScore}{19.8}
\newcommand{\finTopThreeSpread}{4.1}
\newcommand{\hcCostMaxModel}{Claude Fable 5.1 (adaptive/max)}
\newcommand{\rhoDomainsNonzero}{0.82}
\newcommand{\nNonzeroEither}{23}
\newcommand{\finHigherCount}{21}
\newcommand{\hcHigherCount}{2}
\newcommand{\hcHigherNames}{Claude Opus 5.5 (adaptive/max) and Hy3 (high)}
\newcommand{\bothBelowFive}{19}

%% file: generated/tab_dataset.tex
Tasks & \hcTasks & \finTasks & \gdpTasks \\
Request length, words (mean / median) & \hcPromptMean{} / \hcPromptMedian & \finPromptMean{} / \finPromptMedian & \gdpPromptMean{} / \gdpPromptMedian \\
Input files per task (mean / median) & \hcFilesMean{} / \hcFilesMedian & \finFilesMean{} / \finFilesMedian & \gdpFilesMean{} / \gdpFilesMedian \\
Input files per task (range) & \hcFilesMin--\hcFilesMax & \finFilesMin--\finFilesMax & 0--\gdpFilesMax \\
Rubric criteria per task (mean / median) & \hcCritMean{} / \hcCritMedian & \finCritMean{} / \finCritMedian & \gdpCritMean{} / \gdpCritMedian \\
Estimated hours per task (mean)$^\ddagger$ & \hcHoursMean & \finHoursMean$^\S$ & 6.7 \\

%% file: generated/tab_leaderboard.tex
Claude Opus 5.5 (adaptive/max) & Anthropic & 24.7 & 1 & 23.9 & 1 \\
GPT-6 Astra (max) & OpenAI & 11.6 & 2 & 21.5 & 2 \\
Claude Fable 5.1 (adaptive/max) & Anthropic & 9.6 & 3 & 19.8 & 3 \\
Grok 4.7 (xhigh) & xAI & 8.4 & 4 & 14.5 & 5 \\
Muse Spark 1.3 (max) & Meta & 7.8 & 6 & 14.8 & 4 \\
Claude Opus 5 (adaptive/max) & Anthropic & 8.4 & 4 & 11.3 & 7 \\
Grok 4.6 (xhigh) & xAI & 2.8 & 9 & 12.8 & 6 \\
Claude Fable 5 (adaptive/max) & Anthropic & 5.6 & 7 & 9.0 & 9 \\
GPT-6 Sol (max) & OpenAI & 3.6 & 8 & 9.3 & 8 \\
GPT-5.6 Sol (max) & OpenAI & 1.6 & 12 & 5.5 & 11 \\
GLM 5.3 (max) & Zhipu AI & 0.4 & 16 & 6.0 & 10 \\
Qwen 3.8 Max (xhigh) & Alibaba Cloud & 2.0 & 10 & 3.0 & 13 \\
Kimi K3 (max) & Moonshot AI & 1.2 & 14 & 3.8 & 12 \\
Muse Spark 1.2 (xhigh) & Meta & 2.0 & 10 & 2.3 & 16 \\
Gemini 3.8 Flash (high) & Google & 0.8 & 15 & 2.8 & 14 \\
GPT-5.6 Terra (xhigh) & OpenAI & 1.6 & 12 & 1.8 & 20 \\
Gemini 3.7 Flash (high) & Google & 0.0 & 21 & 2.8 & 14 \\
GLM 5.3 Flash (max) & Zhipu AI & 0.4 & 16 & 2.3 & 16 \\
Claude Sonnet 5 (adaptive/max) & Anthropic & 0.0 & 21 & 2.3 & 16 \\
GPT-6 Luna (max) & OpenAI & 0.0 & 21 & 2.3 & 16 \\
GPT-5.6 Luna (max) & OpenAI & 0.4 & 16 & 1.8 & 20 \\
DeepSeek V4 Pro (max) & DeepSeek & 0.4 & 16 & 1.0 & 22 \\
Hy3 (high) & Tencent & 0.4 & 16 & 0.3 & 23 \\
DeepSeek V4 Flash (max) & DeepSeek & 0.0 & 21 & 0.0 & 24 \\
Gemini 3.1 Pro (high) & Google & 0.0 & 21 & 0.0 & 24 \\
Inkling (max) & Thinking Machines & 0.0 & 21 & 0.0 & 24 \\
Kimi K2.7 Code (max) & Moonshot AI & 0.0 & 21 & 0.0 & 24 \\
Mistral Large 3 & Mistral AI & 0.0 & 21 & 0.0 & 24 \\
Muse Glimmer 30B (xhigh) & Meta & 0.0 & 21 & 0.0 & 24 \\
Nemotron 3 Ultra & NVIDIA & 0.0 & 21 & 0.0 & 24 \\